\documentclass[conference,11pt]{IEEEtran}
\IEEEoverridecommandlockouts
\usepackage{cite}
\usepackage{amsmath,amssymb,amsfonts}
\usepackage{algorithmic}
\usepackage{hyperref}
\usepackage{graphicx}
\usepackage{subcaption}
\graphicspath{ {./Figures/} }
\usepackage{tikz}
\usetikzlibrary{shapes.geometric, arrows}
\usepackage{textcomp}
\usepackage{xcolor}
\usepackage[per-mode=fraction]{siunitx}
\def\BibTeX{{\rm B\kern-.05em{\sc i\kern-.025em b}\kern-.08em
    T\kern-.1667em\lower.7ex\hbox{E}\kern-.125emX}}
    
\definecolor{TUMblue}{RGB}{0,101,189}
\definecolor{TUMblue2}{RGB}{0,82,147}
\definecolor{TUMblue3}{RGB}{0,51,89}
\definecolor{TUMwhite}{RGB}{255,255,255}
\definecolor{TUMblack}{RGB}{0,0,0}
\definecolor{TUMgreen}{RGB}{172,163,0}
\definecolor{TUMorange}{RGB}{227,114,34}
\definecolor{TUMlightblue}{RGB}{152,198,234}

\usepackage[utf8]{inputenc}
\usepackage{pgfplots}
\DeclareUnicodeCharacter{2212}{−}
\usepgfplotslibrary{groupplots,dateplot}
\usetikzlibrary{patterns,shapes.arrows}
\pgfplotsset{compat=newest}

\usepackage{pst-node,pst-text,pst-3d}
\usepackage{fancyhdr}

\usepackage{textcomp}
\usepackage{lipsum}
\newcommand\copyrighttext{%
    \footnotesize \textcopyright  2020 IEEE.  Personal use of this material is permitted.  Permission from IEEE must be obtained for all other uses, in any current or future media, including reprinting/republishing this material for advertising or promotional purposes, creating new collective works, for resale or redistribution to servers or lists, or reuse of any copyrighted component of this work in other works.
}
\newcommand\copyrightnotice{%
    \begin{tikzpicture}[remember picture,overlay]
    \node[anchor=south,yshift=1pt, xshift=10pt] at (current page.south) {\fbox{\parbox{\dimexpr\textwidth-\fboxsep-\fboxrule\relax}{\copyrighttext}}};
    \end{tikzpicture}%
}

\begin{document}

\title{{\large \vspace*{-3mm} \hspace*{-3mm} 2020 Fifteenth International Conference on Ecological Vehicles and Renewable Energies (EVER)}\vspace{5mm}\\
Persistent Map Saving for Visual Localization for Autonomous Vehicles: An ORB-SLAM\,2 Extension}

\author{
    \IEEEauthorblockN{
        Felix Nobis\IEEEauthorrefmark{1}, Odysseas Papanikolaou, Johannes Betz and Markus Lienkamp
    }
    \IEEEauthorblockA{
        Chair of Automotive Technology, Technical University of Munich\\
        Munich, Germany \\
        Email: \IEEEauthorrefmark{1}nobis@ftm.mw.tum.de
    }
}

\maketitle
\copyrightnotice


\begin{abstract}
Electric vhicles and autonomous driving dominate current research efforts in the automotive sector. The two topics go hand in hand in terms of enabling safer and more environmentally friendly driving. One fundamental building block of an autonomous vehicle is the ability to build a map of the environment and localize itself on such a map. In this paper, we make use of a stereo camera sensor in order to perceive the environment and create the map. With live Simultaneous Localization and Mapping (SLAM), there is a risk of mislocalization, since no ground truth map is used as a reference and errors accumulate over time. Therefore, we first build up and save a map of visual features of the environment at low driving speeds with our extension to the ORB-SLAM\,2 package. In a second run, we reload the map and then localize on the previously built-up map. Loading and localizing on a previously built map can improve the continuous localization accuracy for autonomous vehicles in comparison to a full SLAM. This map saving feature is missing in the original ORB-SLAM\,2 implementation. 

We evaluate the localization accuracy for scenes of the KITTI dataset against the built up SLAM map. Furthermore, we test the localization on data recorded with our own small scale electric model car. We show that the relative translation error of the localization stays under 1\% for a vehicle travelling at an average longitudinal speed of 36 m/s in a feature-rich environment. The localization mode contributes to a better localization accuracy and lower computational load compared to a full SLAM. The source code of our contribution to the ORB-SLAM2 will be made public at: \href{https://github.com/TUMFTM/orbslam-map-saving-extension}{https://github.com/TUMFTM/orbslam-map-saving-extension}.
\end{abstract}



\IEEEpeerreviewmaketitle

\section{Introduction}

Mastering autonomous driving is an ongoing challenge in current research in the automotive industry and academia. In order to navigate safely, an autonomous vehicle needs to build an accurate representation of its environment and estimate its position within it. In this work, we aim to localize a vehicle along a previously driven and mapped route. It serves as an intermediary development for the application as a localization method in circuit racing scenarios \cite{Betz.2019}. In these conditions, the vehicle needs to be able to localize its position precisely to enable maneuvering at high velocities.

An effective measure for estimating the ego vehicle position is the use of a global navigation satellite system (GNSS). However, the accuracy of these systems is limited. Noise is introduced due to the atmospheric conditions, signal reflections and clock errors. Real Time Kinematic (RTK) positioning systems can reduce these errors by relying on the correction signal of a fixed calibrated base station. On the downside, such systems are reliant on further infrastructure and come with additional costs \cite{InfotipServiceGmbH.2019}.

To be independent of GNSS or RTK coverage for localization, vehicles use lidar, camera and radar sensors to perceive the environment. On the basis of these sensor data, a map representation is computed for the current environment. Simultaneously, the vehicle localizes itself relatively to this map (SLAM). When these local maps are saved at a fixed global GNSS position, vehicles driving in previously mapped locations only need to localize themselves on the given map to get a global position estimation. During a SLAM, a localization error can lead to a discontinuity in the map and thereby hinder any further localization or re-localization. The localization-only mode saves computational resources and enables re-localization in the case of a previous localization error. 

In the following, we adapt the ORB-SLAM\,2 \cite{MurArtal.2017} algorithm to such two-stage localization-only scenarios. In a first step, we create a map of the environment at low speeds through a SLAM. In the second step, we load the map and localize the vehicle on it at higher speeds. We show that the computational load and the localization error can be lowered in this localization-only strategy. We briefly cover related research and review the algorithm we build upon in Section~\ref{sec:related_work}. Section~\ref{sec:implementation} discusses our implementation of the ORB-SLAM\,2 extension. We evaluate our extension in different scenarios in Section~\ref{sec:results} and close the paper with our conclusions.


\section{Related Work}
\label{sec:related_work}
The various onboard sensors for localization - radar, camera and lidar - have individual strengths and drawbacks.

\cite{Ward.2016} present a localization approach on a previously built-up map of radar measurements. The localization error they show is small in terms of lateral positioning, as the features coming from the roadside are distinctive. The error in longitudinal direction is greater since the features are less distinctive in this direction. 

The field of lidar SLAM is widely studied in literature \cite{Grisetti.2007, Kohlbrecher.2011,Hess.2016}. Lidar sensors are widely available in research, especially for indoor robotics. The reliant range measurements give a relatively dense estimation of the environment compared to radar systems. In the open source community, the Adaptive Monte Carlo Localization (AMCL) implementation in the Robot Operating System (ROS) framework is widely used for localization on a 2D occupancy grid, and is applied successfully to a racing scenario by \cite{Stahl.2019}. 

In comparison to radar and lidar data, camera data is more dense and can perceive textures or features on straight walls more easily. It is therefore the sensor with the most potential for localization. Rangan \cite{Rangan.2018} builds up a local map with camera features and associates them with GNSS coordinates. During localization, their system matches the features of the current camera frame against a GNSS-queried subset of the map features. In contrast, we build up a global map which contains all extracted features of the environment. After defining an initial GNSS location for the map, we rely on the stereo camera information only.

\subsection{Visual SLAM}
An extensive overview of current visual SLAM approaches is given in \cite{Saputra.2018}. One difficulty in working with camera sensors for localization purposes is that of depth generation from images. With the knowledge of the camera intrinsic parameters, we can calculate the direction of the light beam that resulted in every pixel in the image. However, the distance from which this light originated and thereby the real world 3D position of the sensed object cannot be measured. In recent years, approaches have been developed to estimate the depth from images of monocular cameras with learning methods \cite{Casser.2018, Alhashim.20190310,Godard.2017}. However, these do not yet possess the same quality as the direct depth calculation from a stereo camera system, which we rely on in this work. 

Two state-of-the-art open source visual SLAM systems are ORB-SLAM\,2\cite{MurArtal.2017} and Direct Sparse Odometry (DSO) \cite{Engel.2018}. Those two systems are representatives of the two main approaches for visual odometry systems: Feature-based methods and direct methods. ORB-SLAM\,2 matches sparse visual features (keypoints and descriptor vectors) between image frames in order to perform the tracking, mapping and place-recognition tasks. DSO minimizes the photometric error of consecutive images in pixel values and does not rely on feature points. Both of these methods have proven their accuracy in the KITTI Vision Benchmark\cite{KittiBenchmark}. Direct methods have gained popularity over the last years due to their high accuracy, efficiency and robustness \cite{Yang.2017}. In this paper, the focus is on evaluating the re-localization accuracy of SLAM systems when a map is already available and pose estimation is challenging, e.g. at high velocities. DSO is a purely visual odometry system which does not perform loop closure and cannot recover its position by re-localizing if tracking is lost. We develop our persistent map saving and loading extension to ORB-SLAM\,2 which has the ability to re-localize the ego vehicle on previously seen frames, using feature extraction and matching.   

\subsection{ORB-SLAM\,2}
In this section, we review some of the workings of the ORB-SLAM\,2 algorithm which are of importance to the proposed extension in this paper. 

Internally, ORB-SLAM\,2 consists of three main parallel threads: Tracking, local mapping and loop closure. The tracking thread localizes the camera with every frame by finding feature matches and minimizing the re-projection error of these features to the previous frame. In parallel, the local mapping thread uses local bundle adjustment to optimize a window of keyframes and keypoints. The loop closing thread is responsible for detecting revisited areas and correcting the accumulated drift using pose-graph optimization with the $g^2o$ framework \cite{g2o}. After the pose-graph optimization, it launches a fourth thread that performs full bundle adjustment and optimizes the whole map and trajectory estimations between keyframes. Global optimization is essential in SLAM systems, as even small trajectory estimation errors accumulate over time, which leads to false trajectory estimates for longer mapping sequences. By performing loop closure and global trajectory optimization, ORB-SLAM\,2 is able to minimize the accumulated error and improve the localization accuracy. 

ORB-SLAM\,2 extracts ORB features \cite{Rublee.2011} from the input images to encode the environment in a sparse representation. These features are robust with regard to rotation and scale, while being fast to extract and match, thus allowing for real-time operation \cite{Karami.2017}. ORB-SLAM\,2 recognizes revisited places based on the DBoW2 library \cite{GalvezLopez.2012}. It compares the current image features in a binary format against the binary features saved in the hierarchical DBoW2 database. If a certain similarity score between the current frame and a representation in the database is reached, a loop closure is detected. The functionality is used to detect loop closures and to perform re-localization in mapped areas in SLAM mode. In localization mode, the local mapping and loop closing threads are deactivated and the system uses only the tracking thread to track the camera in the previously mapped scene. This is accomplished by matching features in the current frame to the features of the previous frame, as well as by mapping points to perform re-localization.


\section{Map Saving Extension Implementation}
\label{sec:implementation}

We use the stereo camera version of the ORB-SLAM\,2 ROS adaption by \cite{JanBrehmer.2019} as the base line for our extension. This section briefly describes the design of the map saving extension and the inner workings of the existing approach which had to be adapted.

Despite the fact that ORB-SLAM\,2 offers the ability to run in localization mode, it does so only in online mode during the same run. The user can set the system to localization mode manually in a Graphical User Interface. However, when the SLAM process ends, the map is deleted. Therefore, we create an extended version of ORB-SLAM\,2 which offers the option of saving the created map in a binary file when the SLAM process ends. After an initial mapping run and a complete system shutdown, the vehicle computer can load the saved map and localize the position of its current camera frame in it. The re-localization is performed via the localization tracking thread described above. Our extension to ORB-SLAM\,2 can easily be integrated as a localization package for robots running on ROS. It can serve as an alternative package to the lidar based AMCL package or fused as an additional input to a more complex localization approach.

\begin{figure}[ht]
\centering
\begin{tikzpicture}[node distance=1.8cm]

    \tikzstyle{startstop} = [rectangle, rounded corners, minimum width=1.5cm, minimum height=0.7cm,text centered, draw=TUMwhite, fill=TUMblue3]

    \tikzstyle{process} = [rectangle, minimum width=1.5cm, minimum height=0.8cm, text centered, text width=2cm, draw=TUMwhite, fill=TUMblue]

    \tikzstyle{decision} = [diamond, minimum width=1cm, minimum height=0.4cm, text centered, text width=1.3cm, draw=TUMwhite, fill=TUMblue2]

    \tikzstyle{arrow} = [thick,->,>=stealth]
    
    \node (start) [startstop, fill=TUMblue3, text=TUMwhite] {\footnotesize{Initialize}};
    \node (dec1) [decision, below of=start, fill=TUMblue2, text=TUMwhite] {\footnotesize{Load Map}};

    \node (proc1) [process, left of=dec1, xshift=-1cm, fill=TUMblue, text=TUMwhite] {\footnotesize{Map Loading}};
    \node (proc2) [process, below of=proc1,yshift=0.5cm, fill=TUMblue, text=TUMwhite] {\footnotesize{Localization Mode}};
    \node (proc4) [process, below of=proc2, fill=TUMblue, text=TUMwhite, yshift=0.6cm] {\footnotesize{Shutdown}};
 
    \node (proc3) [process, right of=dec1, xshift=1cm, yshift=-1.25cm, fill=TUMblue, text=TUMwhite] {\footnotesize{SLAM Mode}};
    \node (proc5) [process, below of=proc3, yshift=0.5cm, fill=TUMblue, text=TUMwhite] {\footnotesize{Shutdown}};
    \node (dec3) [decision, below of =proc5, yshift=-0.2cm, fill=TUMblue2, text=TUMwhite] {\footnotesize{Save Map}};
    \node (proc6) [process, left of=dec3, xshift =-1cm, fill=TUMblue, text=TUMwhite] {\footnotesize{Map Saving}};

    \node (end) [startstop, below of=proc6, yshift=0.1cm, fill=TUMblue3, text=TUMwhite] {\footnotesize{End}};
    
    \draw [arrow] (start) -- (dec1);
    \draw [arrow] (dec1) -- node[anchor=south, xshift= 0.1cm] {yes} (proc1);
    \draw [arrow] (dec1) -| node[anchor=south, xshift=-0.2cm] {no} (proc3);
    \draw [arrow] (proc1) -- (proc2);
    \draw [arrow] (proc2) -- (proc4);
    \draw [arrow] (proc4) |- (end);
    \draw [arrow] (proc3) -- (proc5);
    \draw [arrow] (proc5) -- (dec3);
    \draw [arrow] (dec3) -- node[anchor=south, xshift=0.1cm] {yes} (proc6);
    \draw [arrow] (dec3) |- node[anchor=south, xshift=-0.6cm] {no} (end);
    \draw [arrow] (proc6) -- (end);
\end{tikzpicture}
\caption{Flowchart of the two operating modes: Localization (left) and full SLAM (Right) of ORB-SLAM\,2.}
\label{figure:flowchart}
\end{figure}
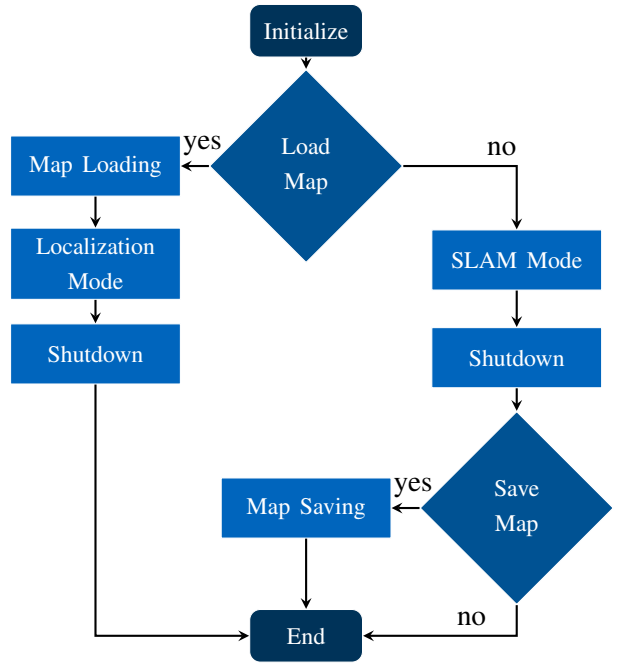

Regarding the implementation details, the localization mode needs to be provided with a map file that contains the objects of the ORB SLAM classes: Map, MapPoint, Keyframe and KeyframeDatabase, and the DBoW2 BoWVector and FeatureVector for each Keyframe. Therefore, we created a SaveMap method that saves this information in a binary file when the SLAM process is finalizing. For consecutive runs, the user can specify via a settings file if he wants the map file to be loaded or not. If the user decides to use a saved map, a LoadMap method is called on system startup which loads the saved Map and KeyframeDatabase and sets the system to localization mode. The latter is done automatically and there is no need for the user to manually change the mode of the system online. Both SaveMap and LoadMap methods are integrated in the System class of ORB-SLAM\,2 which handles all the main functionalities. A high level flowchart of the extended version of ORB-SLAM\,2 can be seen in Figure \ref{figure:flowchart}. 


\section{Experiments and Results}
\label{sec:results}
After introducing the used datasets and metrics in subsection \ref{subsec:dataset_metrics}, we present our method to simulate faster driving speeds using the KITTI dataset in subsection \ref{subsec:kitti_faster_sim}.
The localization evaluation for the KITTI dataset is performed on a powerful desktop CPU in subsection \ref{subsec:kitti}. Additionally experiments on a mobile CPU are presented in subsection \ref{subsec:jetson_eval} for our self-recorded sequences.

\subsection{Datasets and Metrics}
\label{subsec:dataset_metrics}
We evaluate our system using the publicly available KITTI dataset\cite{KittiDataset} and recordings from a model car at 1:10 scale, built during this research project. The model car uses a ZED Stereo camera in order to acquire the visual information about its environment. We run ORB-SLAM\,2 on a platform which uses ROS to test the real-time performance. For the KITTI dataset, we generate ROS bags using the raw data from the Road and City categories which correspond to the Visual Odometry Sequences 01, 07 and 09 recorded on motorway and urban street sections. The ground truth poses are provided and are used for the accuracy evaluation. 

\begin{figure}[ht]
\centering
\includegraphics[width=0.4\textwidth]{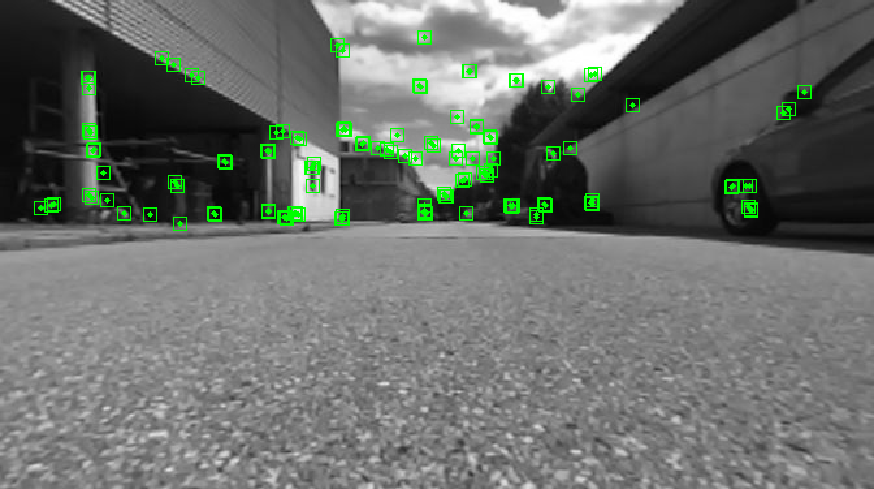}
\caption{Example image frame from the outdoor dataset. ORB feature locations shown in green.}
\label{fig:outdoor}
\end{figure}

\begin{figure}[ht]
	\centering
	\includegraphics[width=0.4\textwidth]{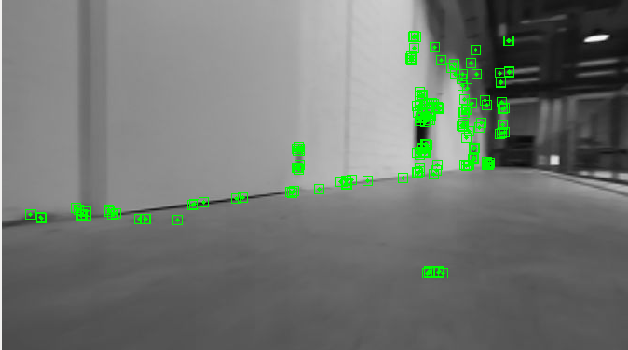}
	\caption{Example image frame from the indoor dataset. ORB feature locations shown in green.}
	\label{fig:indoor}
\end{figure}

We record additional data at an indoor and an outdoor location at the Technical University of Munich with our model car. An example image from the outdoor street setting is shown in Figure \ref{fig:outdoor}. Here, the algorithm is tested in a full scale street environment without traffic. The indoor scene, depicted in Figure \ref{fig:indoor}, is recorded in a the industrial hall of the Chair of Automotive Technology. In this environment, the algorithm has to operate under different lighting conditions and is presented with less distinct optical features due to passages running parallel to straight white walls.

In the data generated with the model car, the paths of the fast driving runs differ from the paths of the slow driving runs which are used for the map creation due to manual driving. This enables us to evaluate our persistent map saving and localization extension in the probable setting where future runs  of an autonomous vehicle will have slightly different trajectories from the mapping trajectory. This adds extra difficulties to the localization due to different perception perspective of the environment.

Before we can perform localization, we build up a map of visual features with our ORB-SLAM\,2 extension. The SLAM mode is computationally demanding. Three threads are running in parallel. In the original implementation, the system reads in a stereo pair of image frames, performs the camera frame tracking and continues with the next pair of frames when the calculations are finished. Our ROS enabled version reads the image frames from a ROS bag at the same rate as they were recorded. This enables us to test the real-time capabilities of the system. The mapping run is performed at moderate driving speeds in order to ensure that ORB-SLAM\,2 does not drop frames and lose track. The localization capability is then evaluated on the same sequences at higher speeds w.r.t. to the initial mapping run.

The metrics used for our evaluation are the percentage of frames where the localization failed (Lost Track: $LT$), the maximum consecutive time that tracking is lost $LT_{t,max}$, the average relative translation $t_{rel}$ and rotation $r_{rel}$ error, as proposed in \cite{KittiDataset}, as well as the absolute translation root-mean-squared error $t_{abs}$ proposed in \cite{TUMBenchmark}. Tracking is lost when ORB-SLAM\,2 fails to perform feature extraction and matching, and is thus unable to track the camera position between consecutive image frames. At high velocities, this can be caused by the higher rotational and translational differences of the camera viewpoints. In those cases, the algorithm can match fewer features, causing an inaccurate localization. In a scene where not many features are observable, the matching might also fail for low speeds. For our own dataset, we resort to the Lost Track metric only as there is no ground truth information available for the trajectory accuracy.

\subsection{High Velocity Simulation for KITTI Data}
\label{subsec:kitti_faster_sim}
The driving speed in the KITTI data is fixed. We simulate higher driving speeds by a higher playback speed of the data. For the simulated speed, non-linearities in the vehicle dynamics might be inaccurately represented. The amount of camera frames available in the simulated dataset is greater than it would be in a recording at faster speeds. To show that the effects of these differences are minor, we record data on our model car with different driving speeds and compare the effect of the playback speed on the results. We record runs at different driving speeds in our dataset. The paths vary slightly due to the manual driving of the model car. We speed up the data of the slow runs to match the same average driving speed in slow and fast runs. The image sequences are then processed by our real-time ORB-SLAM\,2 extension. We measure the percentage of camera frames where the localization fails (Lost Track) in \ref{table:speed_comparison} for those different recordings. We see that the Lost Track metric is similar for runs at high speeds as well as runs with simulated high speeds. As the compared trajectories for the indoor and outdoor category vary in their paths, the slight differences in the metric might be caused by the slightly different amount of features visible in those distinct runs. The comparable results support our choice to simulate faster driving speeds through playback speed variation for the KITTI dataset.

\begin{table}[ht]
\caption{Comparison of fast driving and fast playback speeds on TUM dataset}
\label{table:speed_comparison}
\centering
\begin{tabular}{|c | c | c|} 
 \hline
 Sequence & Average Speed (m/s) & $LT$ (\%) \\ [0.5ex] 
 \hline\hline
 inside\_fast & 6 & 1.5\\
 inside\_slow-4x & 6 & 1.51\\ 
 \hline
 outside\_fast & 7.2 & 0.12  \\ 
 outside\_slow-4x & 7.2 & 0.1  \\
 \hline
 outside\_fast-2x & 14.4 & 4.51  \\
 outside\_slow-8x & 14.4 & 3.97\\ 
 \hline
\end{tabular}
\end{table}

\subsection{Evaluation using the KITTI Dataset}
\label{subsec:kitti}
We evaluate the localization performance using the KITTI Visual Odometry sequences 01, 07 and 09. The algorithm runs on an Intel Xeon CPU E3-1270 v5 and 16 GB of memory. 

\begin{table}[ht]
\caption{SLAM accuracy relative to ground truth}
\label{table:mapping}
\footnotesize
\centering
\begin{tabular}{ |c|c|c|c|c|c|c|  } 
 \hline
 \centering
 Seq. & Speed & Lost Track &  $t_{rel}$ & $r_{rel}$ & $t_{abs}$ \\
 & $(m/s)$ & & $(\%)$ & $(\SI{}{\degree\per100\meter})$ & $(m)$ \\ [0.5ex]
 \hline\hline
 01 & 10 & no & 1.31 &0.2 & 9.89\\  [0.5ex] 
 \hline
 07 & 6 & no & 0.51 & 0.29 & 0.52\\  [0.5ex] 
 \hline
 09 & 8  & no & 0.84 & 0.25 & 1.66\\ 
 09 & 20 & yes  & - & - & -\\  [0.5ex]
\hline
\end{tabular}
\end{table}

First, we need to generate an accurate map, which represents the ground truth trajectory as closely as possible. Table \ref{table:mapping} contains the evaluation results of the mapping runs on the KITTI dataset with respect to the ground truth trajectories. As the localization is performed relative to this map, any errors in the map cannot be corrected by the localization module. Sequence 01 consists of motorway driving with an average linear velocity of $\SI{20}{\meter\per\second}$. As stated in the ORB-SLAM\,2 paper, the translation is difficult to estimate in this sequence because only few feature points close to the vehicle can be tracked. We reduced the playback rate of sequence 01 to 0.5x for the map creation which results in a more precise map. For sequence 09 a playback rate of 0.8x results in the best map accuracy. A further reduction of the playback rate did not improve the accuracy results. At these slower playback speeds, the absolute translational errors for both sequences are lower than the results presented in the original ORB-SLAM\,2 paper.

\begin{figure}[ht]
\centering
\input{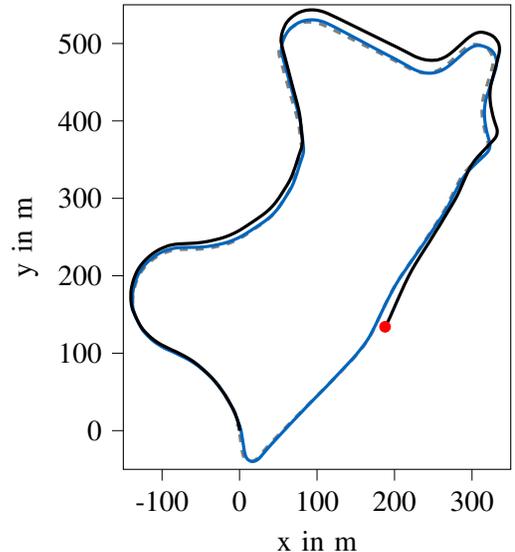}
\caption{\small{KITTI Sequence 09. Ground truth (dashed gray), SLAM trajectories at $\SI{8}{\meter\per\second}$} (blue) and $\SI{20}{\meter\per\second}$ (black). The faster SLAM trajectory diverges further from the ground truth and fails at the position marked in red.}
\label{fig:slam_speed_comparison}
\end{figure}

We also experimented with speeding up of the driving speed for the SLAM runs. Figure \ref{fig:slam_speed_comparison} shows the resulting trajectories for the fast and the slow SLAM runs. For the high speed run, the SLAM mode is no longer able to complete the mapping run, since the feature matching fails. This is marked in red in the figure. In the following, we show that the localization mode is still able to achieve reasonable results at those driving speeds where a SLAM fails.

\begin{table}[ht]
\caption{Localization accuracy relative to map}
\label{table:relativ_localization}
\footnotesize
\centering
\begin{tabular}{ |c|c|c|c|c|c|c| } 
 \hline
 \centering
 Seq. & Speed & $t_{rel}$ & $r_{rel}$ & $t_{abs}$ & $LT$ & $LT_{t,max}$ \\
 & $\SI{}{(\meter\per\second)}$ & $(\%)$ & $(\SI{}{\degree\per100\meter})$ & $\SI{}{(\meter)}$ & $(\%)$ & $\SI{}{(\second)}$ \\ [0.5ex]
 \hline\hline
 01 & 21 &  0.3 & 0.05 & 1.14 & 0 & 0\\  
 01  & 37 &  4.82 & 0.31 & 10.77 & 1.9 & 0.07\\ 
 01  & 42 &  6.12 & 0.41 & 12.82& 6.62 & 0.14\\ [0.5ex] 
 \hline
 07 & 6 &  0.04 & 0.05 & 0.1 & 0& 0\\ 
 07 & 12 &  0.08 & 0.11 & 0.14 & 2.35 & 0.05\\ 
 07 & 18 &  0.23 & 0.13 & 0.22 & 5.87 & 0.49\\ 
 07 & 24 &  0.44 & 0.17 & 0.98 & 8.67 & 0.55\\ [0.5ex] 
 \hline
 09 & 10 & 0.03 & 0.04 & 0.01 & 0 & 0\\ 
 09 & 20 & 0.36 & 0.13 & 0.44 & 0.147 & 0.04 \\ 
 09 & 25 & 0.43 & 0.16 & 0.6 & 2.15 & 0.11\\ 
 09 & 36 & 0.66 & 0.21 & 0.81 & 7.44 & 0.2\\ [0.5ex] 
\hline
\end{tabular}
\end{table}

In the next step, we localize the vehicle relative to the recorded maps at different simulated driving speeds for all sequences. The results are shown in Table \ref{table:relativ_localization}. For each sequence, we replay the trajectory at the same rate as the original recording and increase it by up to 4x the original rate. At the original speed of the recording, the localization algorithm is able to localize the vehicle with errors of only a few centimeters for sequences 07 an 09, while the error on the motorway sequence 01 is higher. 

Furthermore, the results show that a reasonable localization accuracy can be achieved at speeds of up to 36 m/s for sequence 09. The localization approach loses track for about \SI{8}{\percent} of the frames in this case. The longest consecutive time that no localization can be performed is \SI{0.55}{\second}. The presented localization system shall be integrated in a globally fused localization module with additional sensor inputs, so that such a loss rate is acceptable for the use case. The further spatial error metrics increase for higher velocities. For the sequences 07 and 09 these errors stay within a tolerable interval, especially if they are with respect to a longitudinal deviation. A mis-localization along the path is less critical than a lateral deviation in the control concept of the vehicle. For sequence 01, the spatial error metric reaches values which would make a localization impracticable. In this sequence, only few features are present for the feature matching. 

In evaluating the localization algorithm, the relative error with regard to the map is most meaningful. For an application of the overall pipeline the global error with regard to the ground truth is equally important. As this paper focuses on the localization approach, the global error is shown for the sake of completeness in Table \ref{table:global_error} in the appendix.

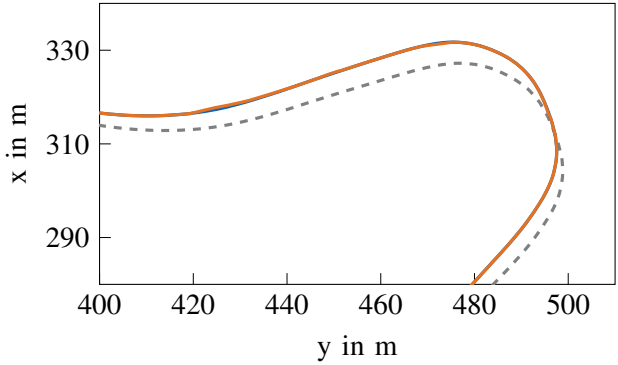
\begin{figure}[ht]
\centering
\begin{tikzpicture}

\definecolor{color0}{rgb}{0,0.396078431372549,0.741176470588235}
\definecolor{color1}{rgb}{0.890196078431372,0.447058823529412,0.133333333333333}

\begin{axis}[
tick align=inside,
tick pos=left,
axis equal image,
x grid style={white!69.01960784313725!black},
xlabel={y in m},
xmin=400, xmax=510,
xtick style={color=black},
xtick={400,420,440,460,480,500},
xticklabels={400,420,440,460,480,500},
y grid style={white!69.01960784313725!black},
ylabel={x in m},
ymin=280, ymax=340,
ytick style={color=black},
ytick={290,310,330},
yticklabels={290,310,330}
]
\addplot [very thick, white!50.19607843137255!black, dashed]
table {%
483.771087646484 279.747894287109
486.882110595703 283.006500244141
488.3623046875 284.60009765625
489.759704589844 286.165496826172
490.428405761719 286.934509277344
491.699493408203 288.479309082031
492.873596191406 290.000213623047
493.947814941406 291.497192382812
494.909393310547 292.943908691406
495.795104980469 294.346099853516
496.606292724609 295.756103515625
497.295501708984 297.157592773438
497.840087890625 298.555908203125
498.080413818359 299.275299072266
498.436187744141 300.703094482422
498.679992675781 302.157806396484
498.775695800781 302.912292480469
498.847290039062 304.383209228516
498.68798828125 306.549499511719
498.413391113281 307.969512939453
497.866394042969 310.082702636719
497.169494628906 312.168487548828
496.634307861328 313.551391601562
496.049011230469 314.930908203125
495.743591308594 315.617492675781
495.063812255859 316.959014892578
493.890502929688 318.877197265625
492.959289550781 320.067810058594
492.437805175781 320.615112304688
491.941589355469 321.170013427734
491.422912597656 321.691802978516
490.286407470703 322.620697021484
488.520111083984 323.795501708984
486.704193115234 324.783599853516
485.450988769531 325.380706787109
484.156188964844 325.901611328125
483.510009765625 326.1455078125
482.161193847656 326.54150390625
481.490509033203 326.712799072266
480.110595703125 326.975006103516
477.981292724609 327.199890136719
476.538513183594 327.226501464844
475.816101074219 327.207611083984
474.330596923828 327.083190917969
472.829498291016 326.886993408203
471.3125 326.612609863281
469.769500732422 326.285888671875
468.194610595703 325.911193847656
466.584686279297 325.484985351562
463.283111572266 324.525787353516
451.547210693359 320.997589111328
445.777313232422 319.1953125
439.958892822266 317.322509765625
436.052703857422 316.159484863281
434.092803955078 315.614501953125
432.119598388672 315.105712890625
430.138610839844 314.633514404297
428.165405273438 314.213592529297
426.182586669922 313.851287841797
424.188812255859 313.546112060547
422.195007324219 313.293701171875
421.191711425781 313.186004638672
419.186309814453 313.026611328125
417.185211181641 312.917999267578
415.179901123047 312.864501953125
413.168212890625 312.853698730469
411.160186767578 312.900390625
409.162109375 312.990386962891
407.162200927734 313.120513916016
405.156707763672 313.297912597656
403.163513183594 313.520294189453
401.177001953125 313.781707763672
399.206512451172 314.075805664062
};
\addplot [very thick, color0]
table {%
478.813842773438 279.232238769531
480.369201660156 280.945129394531
482.707885742188 283.510650634766
485.006195068359 286.019714355469
487.207275390625 288.458312988281
487.909332275391 289.260925292969
489.894104003906 291.645782470703
491.098449707031 293.216186523438
492.202667236328 294.754211425781
492.722900390625 295.500305175781
493.675689697266 296.977172851562
494.128845214844 297.684417724609
494.581665039062 298.432098388672
495.360595703125 299.860260009766
495.718688964844 300.583709716797
496.310363769531 302.01953125
496.585205078125 302.731689453125
496.803558349609 303.444976806641
496.98486328125 304.168426513672
497.294677734375 305.619079589844
497.421356201172 306.391235351562
497.496520996094 307.141479492188
497.547637939453 308.670501708984
497.517150878906 309.409759521484
497.348754882812 310.884887695312
497.244079589844 311.611694335938
497.089813232422 312.326385498047
496.545837402344 314.502197265625
496.080932617188 315.920593261719
495.29150390625 318.048706054688
494.366760253906 320.154846191406
494.042663574219 320.837921142578
493.685760498047 321.519897460938
492.902252197266 322.800415039062
492.478240966797 323.420928955078
491.545440673828 324.632965087891
491.040466308594 325.198883056641
489.980773925781 326.268035888672
488.841186523438 327.219482421875
487.657073974609 328.053039550781
486.435668945312 328.776123046875
485.806884765625 329.1142578125
485.170166015625 329.427062988281
483.873596191406 329.999328613281
483.210540771484 330.260620117188
482.524475097656 330.502136230469
481.139312744141 330.95556640625
480.450286865234 331.129302978516
479.055694580078 331.419921875
478.346252441406 331.522216796875
476.908843994141 331.671569824219
475.465759277344 331.725952148438
474.7353515625 331.721618652344
473.996429443359 331.682006835938
473.2646484375 331.617950439453
471.742645263672 331.435272216797
470.217559814453 331.158538818359
469.436798095703 331.004089355469
467.047729492188 330.435302734375
465.437469482422 330.008544921875
462.971313476562 329.271942138672
458.696655273438 327.945983886719
454.237457275391 326.515014648438
450.558227539062 325.312652587891
445.83447265625 323.740356445312
438.115264892578 321.07958984375
433.281646728516 319.548095703125
431.296203613281 318.952880859375
429.337249755859 318.415191650391
428.347686767578 318.164154052734
427.33056640625 317.884368896484
425.353637695312 317.446533203125
423.363708496094 317.080383300781
422.366943359375 316.908782958984
419.400634765625 316.490905761719
418.401123046875 316.381500244141
414.397918701172 316.080200195312
411.39453125 315.986297607422
409.394195556641 315.992706298828
407.396362304688 316.032012939453
405.412780761719 316.119354248047
404.418609619141 316.184753417969
403.414245605469 316.274383544922
402.415008544922 316.340515136719
400.414794921875 316.543487548828
399.419921875 316.662139892578
};
\addplot [very thick, color1]
table {%
478.034057617188 278.358947753906
481.146484375 281.803009033203
484.266784667969 285.186889648438
486.479858398438 287.630523681641
488.588134765625 290.061248779297
489.894104003906 291.648864746094
491.100006103516 293.214721679688
492.724792480469 295.502014160156
493.674438476562 296.981781005859
494.569976806641 298.4306640625
496.032348632812 301.294616699219
496.576782226562 302.742797851562
496.801208496094 303.439910888672
497.15576171875 304.898162841797
497.406677246094 306.387634277344
497.493835449219 307.149719238281
497.540710449219 307.910522460938
497.559814453125 308.680084228516
497.515808105469 309.411651611328
497.436004638672 310.136352539062
497.302398681641 311.557067871094
496.747589111328 313.791809082031
496.267761230469 315.480407714844
495.816650390625 316.377716064453
495.53857421875 317.111083984375
495.001098632812 318.691711425781
494.686340332031 319.439147949219
494.367950439453 320.133544921875
493.297393798828 322.171691894531
492.903625488281 322.814117431641
492.480010986328 323.439727783203
492.021606445312 324.061614990234
491.537506103516 324.642791748047
490.525939941406 325.754302978516
489.409729003906 326.766387939453
488.839416503906 327.217193603516
486.985290527344 328.526428222656
486.431488037109 328.784912109375
485.803527832031 329.112121582031
483.872863769531 329.996368408203
481.796569824219 330.767211914062
481.127777099609 330.953369140625
479.752960205078 331.283538818359
478.348022460938 331.519836425781
476.911315917969 331.669036865234
475.465881347656 331.730682373047
468.626434326172 330.826995849609
466.226806640625 330.220031738281
464.621459960938 329.763244628906
460.431365966797 328.489685058594
459.144256591797 328.065277099609
457.448516845703 327.555419921875
455.662109375 326.989410400391
453.327209472656 326.222473144531
451.49169921875 325.624847412109
450.697418212891 325.464050292969
448.800354003906 324.836059570312
447.761535644531 324.461608886719
446.720031738281 324.044647216797
442.900909423828 322.709503173828
441.601318359375 322.272979736328
435.811279296875 320.438018798828
431.873596191406 319.269775390625
429.862060546875 318.783752441406
428.723510742188 318.552886962891
427.732971191406 318.321075439453
425.761474609375 317.938812255859
424.764556884766 317.75830078125
419.398132324219 316.491790771484
417.396667480469 316.288970947266
413.392578125 316.039031982422
412.384307861328 315.99951171875
409.421966552734 315.979217529297
406.401733398438 316.064544677734
405.409423828125 316.119903564453
403.4228515625 316.257995605469
401.412475585938 316.436889648438
399.423950195312 316.654357910156
};

\end{axis}

\end{tikzpicture}
\caption{KITTI sequence 09. Ground truth (dashed gray), SLAM trajectory at $\SI{8}{\meter\per\second}$ (blue) and localization at $\SI{36}{\meter\per\second}$ (orange). The SLAM trajectory has an error toward ground truth. Localization aligns well with the SLAM trajectory.}
\label{fig:low_loc_high_map_error}
\end{figure}

\begin{figure}[ht]
\centering
\input{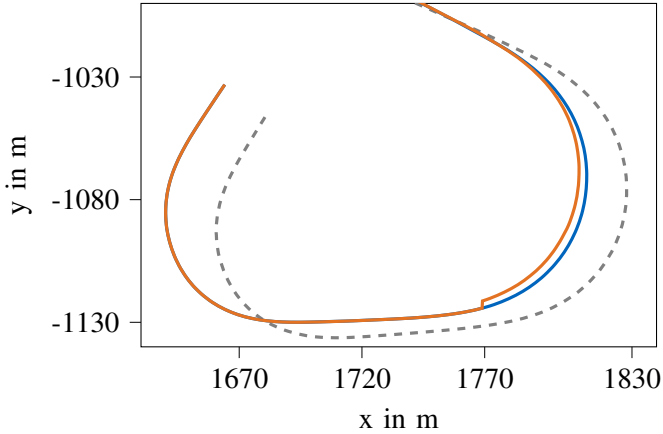}
\caption{KITTI sequence 01. Ground truth (dashed gray), SLAM trajectory at $\SI{10}{\meter\per\second}$ (blue) and localization at $\SI{42}{\meter\per\second}$ (orange). The SLAM trajectory has an error toward ground truth. Localization has an additional error towards the SLAM trajectory for a section of the curve.}
\label{fig:loc_fail_sequence_one}
\end{figure}

Various qualities of the mapping and localization algorithm are shown in the exemplary figures below. Figure \ref{fig:low_error_curve} shows a curve where the mapping and localization results are in close alignment to the ground truth. The blue SLAM trajectory is mostly covered by the localization trajectory in the figure as their paths coincide greatly. Figure \ref{fig:low_loc_high_map_error} shows the localization in close alignment to the reference SLAM trajectory, and confirms the good result of the localization module in sequence 09. The global error however remains significant, as the mapping mode is not able to create an accurate global map. Figure \ref{fig:loc_fail_sequence_one} shows estimated trajectories for sequence 01 where both the mapping and the localization perform poorly due to a lack of features. Generally, it is observed that the system introduces the greatest localization errors during turns with high angular velocities. Translational movements or slow turns introduce less noise. We assume this is due to the greater feature similarity in the latter cases.

\begin{figure}[ht]
\centering
\begin{tikzpicture}

\definecolor{color0}{rgb}{0,0.396078431372549,0.741176470588235}
\definecolor{color1}{rgb}{0.890196078431372,0.447058823529412,0.133333333333333}

\begin{axis}[
tick align=outside,
tick pos=left,
width=8.7cm,
axis equal image,
x grid style={lightgray!92.02614379084967!black},
xlabel={x in m},
xmin=-100, xmax=90,
xtick style={color=black},
xtick={-100,-50,0,50,100},
xticklabels={-100,-50,0,50,100},
y grid style={lightgray!92.02614379084967!black},
ylabel={y in m},
ymin=220, ymax=350,
ytick style={color=black},
ytick={200,250,300,350},
yticklabels={200,250,300,350},
y label style={yshift=-.5em}
]
\addplot [very thick, lightgray!66.92810457516339!black, dashed]
table {%
-101.070899963379 227.792694091797
-98.5490798950195 228.934799194336
-96.0257415771484 229.93620300293
-93.5116729736328 230.819198608398
-91.0401229858398 231.572906494141
-88.5774612426758 232.223403930664
-86.138427734375 232.764205932617
-83.7128219604492 233.198806762695
-81.317138671875 233.531600952148
-78.9529800415039 233.773895263672
-76.5971374511719 233.928604125977
-73.2217102050781 234.046997070312
-68.8815383911133 234.098693847656
-58.5896186828613 234.109497070312
-54.6312713623047 234.221298217773
-51.6917991638184 234.380798339844
-47.8287315368652 234.685104370117
-44.0087814331055 235.101196289062
-40.2265586853027 235.615005493164
-37.4365081787109 236.072296142578
-32.8112106323242 236.970901489258
-27.1205196380615 238.152297973633
-23.2884006500244 239.014602661133
-20.4201507568359 239.73779296875
-17.5858993530273 240.557907104492
-14.7877397537231 241.484298706055
-12.0277795791626 242.497802734375
-9.29703426361084 243.589797973633
-6.61586809158325 244.740997314453
-4.00088500976562 245.9501953125
-1.4451709985733 247.223495483398
1.04535305500031 248.560806274414
3.47195196151733 249.950805664062
5.86499404907227 251.410293579102
8.2351245880127 252.947799682617
12.1562299728394 255.643203735352
29.1395092010498 267.067291259766
31.5153503417969 268.786987304688
34.6061592102051 271.143707275391
36.8851699829102 273.027099609375
39.0970497131348 274.998504638672
41.2907981872559 277.096801757812
43.4392509460449 279.297302246094
44.8305397033691 280.833709716797
46.8570404052734 283.240203857422
48.8008689880371 285.763305664062
50.6780586242676 288.382690429688
52.5037612915039 291.091613769531
54.8752288818359 294.809692382812
62.0438613891602 306.330596923828
63.7611503601074 309.296203613281
64.8348693847656 311.301391601562
66.3361892700195 314.342987060547
67.26171875 316.392913818359
68.116081237793 318.458190917969
69.2966232299805 321.57958984375
71.1851272583008 326.783386230469
72.2398376464844 329.914489746094
73.2254104614258 333.066009521484
76.2866897583008 343.179290771484
78.8046417236328 350.385894775391
};

\addplot [very thick, color0]
table {%
-101.072570800781 230.233444213867
-98.4714202880859 231.402954101562
-95.8889617919922 232.409866333008
-93.3202209472656 233.282989501953
-90.7518920898438 234.071685791016
-88.2017593383789 234.745986938477
-85.6759490966797 235.302795410156
-83.1713790893555 235.752304077148
-80.6840362548828 236.103546142578
-78.2200088500977 236.347259521484
-75.8135070800781 236.507629394531
-72.3048248291016 236.625732421875
-65.6162567138672 236.67561340332
-60.3182907104492 236.669540405273
-56.2349510192871 236.712280273438
-49.2726440429688 237.058532714844
-45.4038391113281 237.375747680664
-41.563591003418 237.81770324707
-36.8449363708496 238.489379882812
-31.2960186004639 239.540313720703
-25.5693531036377 240.712341308594
-21.7502975463867 241.594192504883
-18.9047927856445 242.334533691406
-16.0777015686035 243.156936645508
-13.2640762329102 244.08903503418
-10.4908828735352 245.095809936523
-8.65525817871094 245.810974121094
-5.08970642089844 247.324157714844
-1.62044525146484 248.934127807617
1.73519897460938 250.643264770508
4.99993133544922 252.487411499023
10.6037750244141 255.938461303711
17.8091506958008 260.566711425781
26.757568359375 266.467529296875
30.8007354736328 269.232177734375
33.141242980957 270.939331054688
35.4513092041016 272.6962890625
36.959846496582 273.908782958984
39.2064590454102 275.806701660156
41.3716430664062 277.803039550781
42.8333129882812 279.210296630859
44.2668151855469 280.654510498047
46.3483581542969 282.915130615234
48.3668212890625 285.285461425781
49.6733932495117 286.926696777344
51.5549468994141 289.492797851562
52.8006896972656 291.260559082031
54.6247329711914 293.9814453125
56.4215927124023 296.784362792969
63.0905838012695 307.308288574219
64.8513870239258 310.250305175781
66.4889221191406 313.262451171875
68.0117721557617 316.293334960938
69.3885192871094 319.393737792969
70.2279815673828 321.455505371094
73.3007659912109 329.828430175781
75.0128479003906 335.105560302734
77.4860000610352 343.343200683594
78.483024597168 346.197113037109
80.1306457519531 350.680419921875
};
\addplot [very thick, color1]
table {%
-101.073699951172 230.23616027832
-99.7732009887695 230.835906982422
-95.888671875 232.412673950195
-94.6059417724609 232.869995117188
-90.7432250976562 234.075408935547
-88.2072448730469 234.757507324219
-86.7493591308594 234.931213378906
-85.624153137207 235.282440185547
-84.0937728881836 236.247024536133
-82.8284912109375 236.48469543457
-81.6144332885742 236.629043579102
-79.2125549316406 237.011032104492
-77.0083847045898 236.44172668457
-75.824089050293 236.513092041016
-73.9174652099609 237.070953369141
-70.2914810180664 236.947555541992
-67.8690032958984 236.73942565918
-61.354377746582 236.663360595703
-57.2567977905273 236.705093383789
-54.2377090454102 236.787078857422
-51.2286415100098 236.932174682617
-46.3530120849609 237.294204711914
-44.4295959472656 237.475936889648
-39.675479888916 238.068023681641
-36.8388595581055 238.500442504883
-34.9875602722168 238.825103759766
-31.2987403869629 239.536880493164
-29.2634353637695 240.419692993164
-29.1802673339844 240.080291748047
-28.443338394165 240.099792480469
-22.706485748291 241.361282348633
-19.8610496520996 242.07942199707
-17.0215644836426 242.870178222656
-13.2632331848145 244.090209960938
-10.5035171508789 245.090209960938
-7.74259185791016 246.192764282227
-3.34629821777344 248.083831787109
0.0575408935546875 249.773895263672
3.36781311035156 251.557952880859
5.02395629882812 252.493957519531
8.18723297119141 254.420257568359
15.3852996826172 258.997833251953
25.127685546875 265.386657714844
29.9866485595703 268.675842285156
33.9149780273438 271.521209716797
35.3370132446289 272.455810546875
37.6561813354492 274.374694824219
39.1384582519531 275.660705566406
40.6455917358398 277.125946044922
42.1047210693359 278.498870849609
44.2681045532227 280.658752441406
44.9713821411133 281.397827148438
46.3462677001953 282.915496826172
48.3644256591797 285.316101074219
49.6728286743164 286.93896484375
52.1766891479492 290.367919921875
53.10302734375 291.998474121094
53.5191879272461 292.412017822266
54.0308609008789 293.06591796875
55.9068145751953 295.934600830078
57.9137878417969 298.945922851562
59.7205047607422 301.787780761719
61.0710983276367 303.634368896484
61.879035949707 305.343078613281
64.848762512207 310.246551513672
65.9523162841797 312.250762939453
67.0066528320312 314.270629882812
67.7932891845703 315.446166992188
68.7258605957031 317.509765625
69.5823364257812 319.585510253906
69.990348815918 320.639190673828
71.0055923461914 323.515563964844
72.5694885253906 327.747161865234
74.0083999633789 331.917419433594
75.3272094726562 336.174987792969
77.818359375 344.326934814453
78.7092971801758 346.278167724609
79.2012329101562 347.721466064453
79.8719177246094 349.327880859375
80.5322494506836 351.069061279297
};
\end{axis}

\end{tikzpicture}
\caption{KITTI sequence 09. Ground truth (dashed gray), SLAM trajectory at $\SI{8}{\meter\per\second}$ (blue) and localization at $\SI{36}{\meter\per\second}$ (orange). Both SLAM and localization align well with the ground truth.}
\label{fig:low_error_curve}
\end{figure}
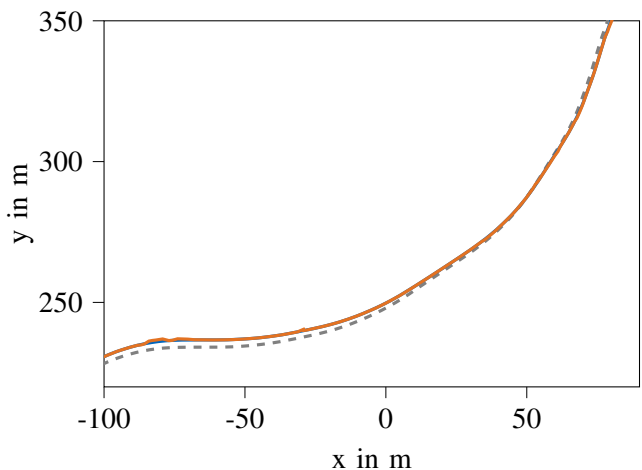

\subsection{Evaluation using the Self-recorded Datasets}
\label{subsec:jetson_eval}
Following the localization evaluation for the KITTI dataset, we evaluate the performance using our self-recorded indoor and outdoor dataset. The trajectories at different speeds vary slightly, making for a more realistic replication of the practical use of the implementation.

In mobile robotics, whether it is self-driving cars or model cars, computational resources are a limitation factor which has to be considered. In addition to the evaluation on the hardware used in subsection \ref{subsec:kitti},  we evaluate the model car localization on the mobile NVIDIA Jetson TX2. The CPU cluster of the Jetson TX2 consists of a dual-core Denver 2 processor and a quad-core ARM Cortex-A57. In this way, we investigate the extent to which localization accuracy is dependent on available computing power.  

\begin{figure}[ht]
\centering
\input{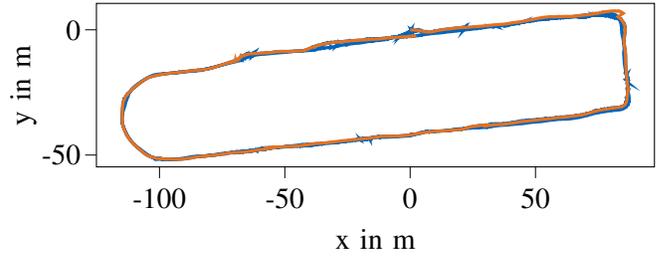}
\caption{TUM Outdoor scene. SLAM trajectory (blue) and localization (orange). Localization and SLAM align well.}
\label{fig:outdoor_traj}
\end{figure}

\begin{figure}[ht]
\centering
\input{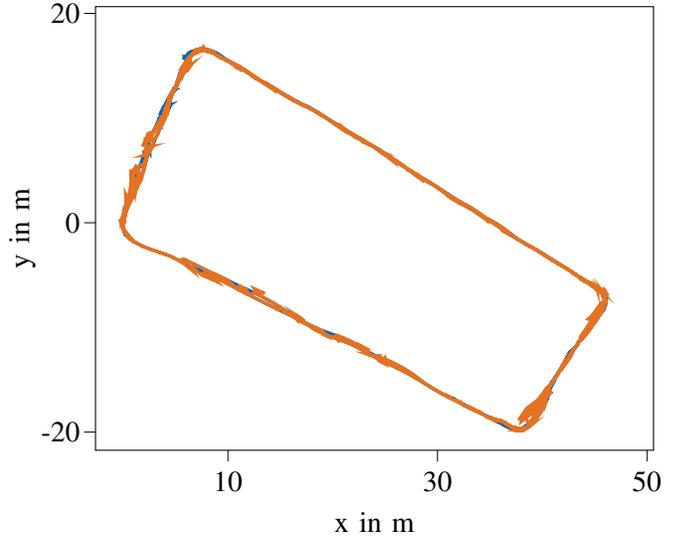}
\caption{TUM Indoor scene. SLAM trajectory (blue) and localization (orange). Localization and SLAM align well. We can see a drift in both trajectories as they do not form a rectangular shape, even though the recorded track is rectangular.}
\label{fig:indoor_traj}
\end{figure}

The SLAM mode is run using our dataset in order to create the initial maps. The trajectories of indoor and outdoors scenes are shown in Figures \ref{fig:outdoor_traj} and \ref{fig:indoor_traj}. For the localization evaluation, only the Lost Track metric is considered since no ground truth is available.

The localization is evaluated for different speeds and trajectories against the SLAM map recorded at low speed. The localization aligns well with the SLAM trajectory for the fast computational hardware which are examplatory shown in Figures \ref{fig:outdoor_traj} and \ref{fig:indoor_traj}. The results for the localization runs on the different computational devices are shown in Table \ref{table:local_dataset}.

\small
\begin{table}[ht]
	\caption{Localization accuracy with different computational hardware}
	\label{table:local_dataset}
	\footnotesize
	\centering
	\begin{tabular}{ |c|c|c|c|c|c|  } 
		\hline
		
		\multicolumn{2}{|c|}{} & \multicolumn{2}{|c|}{Intel Xeon} & \multicolumn{2}{|c|}{Jetson TX2}\\
		\hline
		\centering
		Sequence  & \hspace{-0.3cm}Speed \hspace{-0.3cm} & $LT$ &\hspace{-0.1cm} $LT_{t,max}$ \hspace{-0.2cm} & $LT$ & \hspace{-0.2cm}$LT_{t,max}$\hspace{-0.2cm}\\
		& $(m/s)$ & $(\%)$ & $\SI{}{(\second)}$ & $(\%)$ & $\SI{}{(\second)}$ \\ [0.5ex]
		\hline\hline
		inside\_slow-3x  & 4.5 & 0.9 & 0.21 & 6.93& 0.65\\ 
		inside\_fast  & 6 & 1.5& 0.31 & 10.33 & 0.54\\ 
		\hline
		outside\_slow-2x & 3.6 & 0.07& 0.04 & 7.25& 1.68\\ 
		outside\_fast & 7.2 & 0.12 & 0.02 & 19.23 &1.82\\ 
		\hline
	\end{tabular}
\end{table}

\normalsize
For the Xeon CPU, the amount of frames lost is considerably lower than the one for the NVIDIA Jetson. The embedded CPUs do not have the necessary processing power to perform the necessary feature extraction and matching in real-time. This leads to a loss of tracking - not due to a lack of features, but due to the real-time constraints.

On the Xeon processor the computational resources did not limit the algorithm. Here, we notice that in the indoor environment, the localization error is greater than for the outdoor scenes. This is presumably caused by the fact that the indoor environment includes areas with texture-less white walls which possess few features to be matched. 

More significantly, we see that the matching fails for even lower speeds when comparing it to the results using the KITTI dataset in Section \ref{table:relativ_localization}. There are discontinuities in the estimated trajectories which can be identified as spikes in Figures \ref{fig:outdoor_traj} and \ref{fig:indoor_traj}. The camera images of the model car are taken at a resolution of 672x376 pixels compared to the 1241x376 resolution of the KITTI dataset. Furthermore, the model car uses an input in a JPG-compressed format. The artifacts resulting from the compression are an additional challenge for the feature matching module. The necessity of good image quality for the successful implementation of the localization module is shown here.


\section{Conclusions}
\label{sec:conlusions}
This work presents an extension to a visual SLAM system via a map saving functionality. Like the original implementation, this extension enables the system to be used in slow driving conditions as a SLAM module. Additionally, it can be used in faster driving conditions as a localization module resulting in an overall more precise trajectory estimation. The system is an extension to ORB-SLAM\,2 and can be used for localization in both outdoor and indoor environments. Experimental results show that it is possible to perform localization at velocities of 36 m/s in feature-rich environments.

The localization module requires fewer computational resources than the SLAM module; however, the feature calculation and matching is demanding. We showed the limitations of the approach on an embedded CPU. 

The localization does not perform well for environments with few features, such as KITTI sequence 01. This poses an open challenge to the application of such algorithms to racing scenarios where similar conditions apply. Emerging approaches use features generated by deep learning techniques \cite{DeTone.20170724} for SLAM applications. It is still an open research question as to whether this approach can increase the localization accuracy for the presented scenarios. At the same time, these deep features can be less variant to changing lighting conditions compared to the current optical features. This could make a re-localization possible also for different conditions such as day and night time.

The map saving extension developed and evaluated in this work improves the functionality of the original ORB-SLAM\,2 approach and enables new use cases such as the application to autonomous racing for model cars.

\section*{Contributions and Acknowledgements}
\label{sec:contributions}
Felix Nobis initiated the idea of this paper and contributed essentially to its conception and content. Odysseas Papanikolaou contributed to the implementation and experimental results of this research. Johannes Betz revised the paper critically. Markus Lienkamp made an essential contribution to the conception of the research project. He revised the paper critically for important intellectual content. He gave final approval of the version to be published and agrees to all aspects of the work.
As a guarantor, he accepts the responsibility for the overall integrity of the paper. We express gratitude to Continental Engineering Service for funding for the underlying research project.

\bibliographystyle{./bibliography/IEEEtran} 
\bibliography{./bibliography/bibliography}

\newpage
\appendix

\section{Global Localization Error}
In the paper we presented the relative error of the localization module to the SLAM map. The global localization errors towards the ground truth are in general higher than the relative errors, due to addition of the errors arising during the map creation. The global error is presented in Table \ref{table:global_error}. To reduce the global error further, a more accurate map needs to be generated in the SLAM process. This could be achieved by fusing additional sensor data during the SLAM to stabilize the trajectory estimate e.g. through the use of GNSS systems. This is an approach widely considered in the automotive industry, given that a variety of sensor systems is fused for autonomous driving tasks\cite{Marti.2019}.

\begin{table}[ht]
\caption{Localization accuracy compared to ground truth}
\label{table:global_error}
\footnotesize
\centering
\begin{tabular}{ |c|c|c|c|c|c|c|  } 
 \hline
 \centering
 Seq. & Mode & Speed & $LT$ &  $t_{rel}$ & $r_{rel}$ & $t_{abs}$ \\
 & & $(m/s)$ & $(\%)$ & $(\%)$ & $(\SI{}{\degree\per100\meter})$ & $(m)$ \\ [0.5ex]
 \hline\hline
 01 & SLAM & 10 & 0 & 1.31 &0.2 & 9.89\\ 
 01 & LOC  & 37 & 1.9 & 6.36 & 0.82 & 13.89\\ 
 01 & LOC  & 42 & 6.62 & 9.51 & 0.92 & 15.24\\ [0.5ex] 
 \hline
 07 & SLAM & 6 & 0 & 0.51 & 0.29 & 0.52\\ 
 07 & LOC  & 12 & 2.35 & 0.62 & 0.33 & 0.53\\ 
 07 & LOC  & 18 & 5.87 & 0.7 & 0.37 & 0.56\\ 
 07 & LOC  & 24 & 8.67 & 0.93 & 0.42 & 1.12\\ [0.5ex] 
 \hline
 09 & SLAM & 8  & 0 & 0.84 & 0.25 & 1.66\\ 
 09 & SLAM & 20 & Fails  & - & - & -\\ 
 09 & LOC  & 20 & 0.147 & 1.01 & 0.53 & 1.71 \\ 
 09 & LOC  & 25 & 2.15 & 1.09 & 0.55 & 1.72\\ 
 09 & LOC  & 36 & 7.44 & 1.27 & 0.58 & 1.8\\ [0.5ex]
\hline
\end{tabular}
\end{table}

\end{document}